\documentclass[conference]{IEEEtran}
\IEEEoverridecommandlockouts
\usepackage{cite}
\usepackage[dvipsnames]{xcolor}   % for named colors
\usepackage[colorlinks=true,
            linkcolor=MidnightBlue,   % section refs, equation refs
            citecolor=NavyBlue,       % \cite
            urlcolor=BrickRed]        % \url, \href
           {hyperref}
\usepackage{amsmath,amssymb,amsfonts}
\usepackage{algorithmic}
\usepackage{graphicx}
\usepackage[export]{adjustbox}
\usepackage{textcomp}
\usepackage{xcolor}
\usepackage{caption}
\usepackage{multirow, multicol}
\usepackage{nameref}
\usepackage{booktabs,threeparttable}
\usepackage[font=small,skip=2pt]{caption}
\usepackage{subcaption}
\usepackage{float}
\DeclareMathAlphabet{\mathcal}{OMS}{cmsy}{m}{n}
\SetMathAlphabet{\mathcal}{bold}{OMS}{cmsy}{b}{n}
\usepackage{hyperref}
\def\BibTeX{{\rm B\kern-.05em{\sc i\kern-.025em b}\kern-.08em
    T\kern-.1667em\lower.7ex\hbox{E}\kern-.125emX}}
\begin{document}

\title{
Point Virtual Transformer
}

\author{
\IEEEauthorblockN{Veerain Sood}
\IEEEauthorblockA{Computer Science \& Engineering, \\
IIT, Tirupati, India \\
Email: veerainsood1@gmail.com}
\and
\IEEEauthorblockN{Bnalin}
\IEEEauthorblockA{Texas A \& M University\\
Email: bnalin@tamu.edu}
\and
\IEEEauthorblockN{Gaurav Pandey}
\IEEEauthorblockA{Engineering Technology \& Industrial Distribution \\
Texas A \& M University\\
Texas, USA \\
gpandey@tamu.edu}
}
\maketitle

\begin{abstract}
LiDAR-based 3D object detectors often struggle to detect far-field objects due to the sparsity of point clouds at long distances. The limited number of points corresponding to objects at such long ranges fails to provide concrete geometrical cues needed for reliable object detection. To mitigate this issue, depth-completion techniques are used to generate `virtual points' from RGB images. However, using all virtual points directly can lead to excessive computational overhead, making the use of selective sampling or voxelization essential. Additionally, selecting the right fusion strategy between real and virtual data is crucial for optimal performance. In this paper, we propose the \textbf{Point Virtual Transformer (PointViT)}, a transformer-based detector that is designed to reason jointly on raw and virtual LiDAR data. We investigate multiple fusion strategies, including early fusion to BEV-based gated fusion techniques, and analyse their relative advantages and trade-offs. At a high level, the approach first fuses the original point cloud with long-range virtual points using one of several explored fusion mechanisms. The resultant hybrid point cloud is voxelized and encoded via a series of sparse submanifold convolutions to generate a BEV heatmap that highlights potential object centres. A query initialisation module then identifies a compact, coverage-aware set of high-confidence candidate queries from the BEV heatmap through score-modulated farthest point sampling. Each selected query is projected into the 3D space using a lightweight vote head, which estimates offsets between potential object centres and the current query coordinates. Finally, these heatmap-aligned queries interact with local voxel and point features through a transformer-based context assembly module, enabling robust reasoning across near-field and far-field regions. The proposed model achieves 91.16\% 3D AP and 95.94\% BEV AP on the KITTI 3D object detection dataset, as well as 99.36\% on the 2D detection benchmark (Car class) on KITTI 2D object detection.
\hyperlink{https://github.com/Veerainsood/PointVit}{https://github.com/Veerainsood/PointVit}

\end{abstract}
\section{Introduction}

The prevalence of autonomous systems in the past decade has led to a rise in the development of innovative 3D object detection methods~\cite{b2, b3, b4, b6}. Although LiDAR sensors provide reliable depth measurements, the resulting point clouds are typically sparse, especially for distant or partially occluded objects.  This sparsity reduces the detector’s ability to accurately detect objects at long ranges. Camera images, on the other hand, possess rich semantic information based on colour and texture that is often absent in LiDAR data. 
Combining the two sources of information to produce a stronger detector has become a significant topic of research~\cite{b3}. However, this process remains challenging due to the difference in sampling densities and the computational burden introduced by dense projections. Recent work~\cite{b36} suggests the use of \emph{virtual points} (i.e., the points generated by depth completion techniques which are then back-projected into the LiDAR plane) to enhance the point cloud by filling up the far-field sparsity of objects, giving better geometrical cues for the model to detect objects at those locations.
\begin{figure}[H]
  \centering
  \begin{subfigure}[t]{0.49\linewidth}
    \centering
    \includegraphics[height=5.0cm,keepaspectratio]{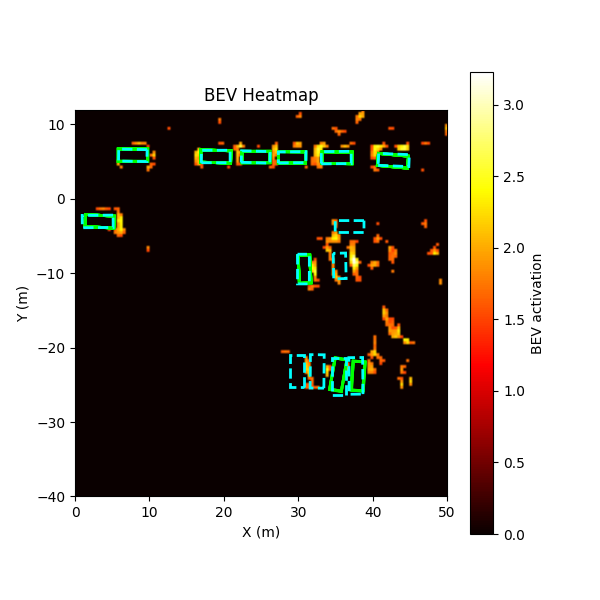}
    \caption{}
  \end{subfigure}\hfill
  \begin{subfigure}[t]{0.49\linewidth}
    \centering
    \includegraphics[height=5.0cm,keepaspectratio]{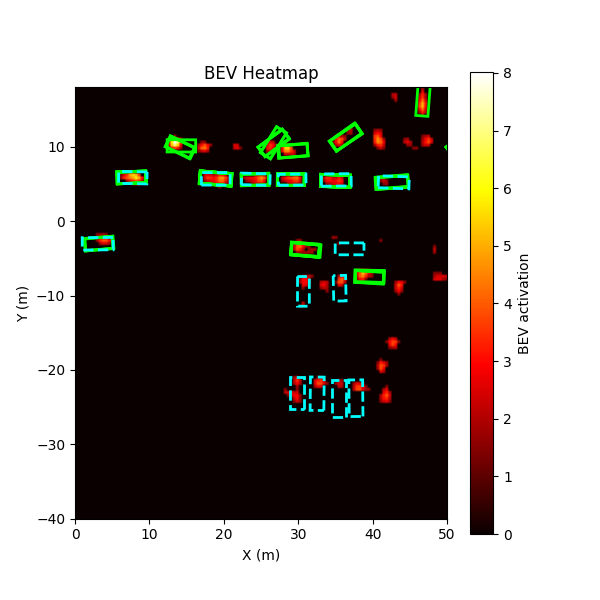}
    \caption{}
  \end{subfigure}
  \caption{BEV heatmaps generated using (a) LiDAR points fused with Virtual Points, and (b) raw LiDAR points only. 
In (a), the bottom region exhibits stronger activations, indicating enhanced feature representation and denser coverage in previously sparse areas. 
This improvement allows the model to detect objects more reliably and with greater spatial consistency. 
\textcolor{Green}{Green boxes} denote \textcolor{Green}{correctly} detected objects, \textcolor{Cyan}{Cyan boxes} indicate \textcolor{Red}{missed} detections, and purple boxes represent additional detections enabled by the introduction of Virtual Points.}
\label{fig:fig1}
\end{figure} 
However, the number of \emph{virtual points} is often orders of magnitude larger than the number of points in the original point cloud. Hence, simply adding them to the original point cloud results in increased computation with little gain. Here, we propose an efficient point virtual transformer with various fusion strategies that enhance far-field object detection without requiring the inclusion of all virtual points. The key contributions of our work can be summarised as follows:
\begin{itemize}
    \item We propose a unified framework that complements state-of-the-art transformer-based 3D object detectors with virtual-point-based cross-modal fusion strategies to improve long-range detection.

    \item We compare the performance of three different cross-modal fusion strategies, namely Early Fusion, Late Gated Fusion, and Late Convolution Fusion, within our unified framework as summarised in Table \ref{tab:ablation-summary}.

    %\item We study the effects of using different types of virtual point generators on the same underlying architecture to investigate the impact on the performance and accuracy of the resulting model.
    % \item We study the effect of different virtual point generation methods, applying them within a transformer framework, and quantify their impact on 3D object detection at long ranges.
    
    \item We demonstrate that virtual point fusion leads to enhancement of 2D object detection, yielding strong results in all difficulty levels as seen in Table \ref{tab:kitti-2d-curated}
    
\end{itemize}

\section{Related Work}
\subsection{\textbf{LiDAR-Only Object Detection}}
Object detectors like VoxelNet~\cite{b45} and SECOND~\cite{b22} represent 3D space as voxel grids and rely on voxel features to balance efficiency and accuracy of object detection.
PointPillars~\cite{b23}, on the other hand, compresses voxel features along the vertical axis, which improves inference speed. Models like PV-RCNN~\cite{b13,b17} and Voxel-RCNN~\cite{b3} combine voxel and point features, and models like PDV~\cite{b16} rely on point cloud density to refine voxel feature representations.
The main drawback of these models is that their performance drops significantly for distant objects where LiDAR sampling is too sparse to describe fine geometry.

\subsection{\textbf{Multi-Modal and Virtual-Point Fusion}}
To mitigate the limitations of LiDAR-only object detection, models like PointPainting~\cite{b3} perform semantic segmentation and attach segmented labels to the corresponding LiDAR points, whereas MVX-Net~\cite{b2} fuses image-level features with LiDAR points at the voxel-level. Such approaches do enrich semantics but fail to address the point sparsity problem for distant regions.
Depth completion models, such as \cite{b46,b47,b39,b36}, generate pseudo or virtual points from camera images to create dense 3D representations.
This idea addresses the distant-region sparsity problem, but increases computational cost. Moreover, this method also adds noise, which decreases the detection accuracy. ViKIENet~\cite{b32} preferentially selects semantically rich regions and projects those regions into the LiDAR plane after aligning with raw LiDAR features using attention mechanisms. This selective approach reduces the computational overhead associated with virtual points, but also removes subtle details of objects and discards direct colour information. %Motivated by~\cite{b36} we removed a large portion of redundant nearby virtual points and retained a denser set of far-field color-augmented virtual points. We then employ attention mechanisms to learn the structure of far-field sparse objects using virtual and real LiDAR points. This results in a representation that preserves more spatial coverage and adds syntactic clues for the model, enabling better detection of distant objects.

\subsection{\textbf{Transformer-Based 3D Detection}}
The ability of the attention mechanism to add contextual awareness in sequential data has been intensively used after it was first showcased in~\cite{b5}, eventually leading to the application of transformers to computer vision, including tasks like 3D object detection. Methods such as Voxel Transformer~\cite{b20}, M3DETR~\cite{b14}, and PVT-SSD~\cite{b6} utilise attention mechanisms to capture long-range structural dependencies in the point cloud.
PVT-SSD~\cite{b6} also employs a query initialization module, which selects object-centric candidates from BEV heatmaps using a score-modulated farthest-point sampling strategy, thus achieving good spatial coverage and reducing computational costs.

\section{Method}
\begin{figure*}[!t]
  \centering
  \includegraphics[width=.9\textwidth]{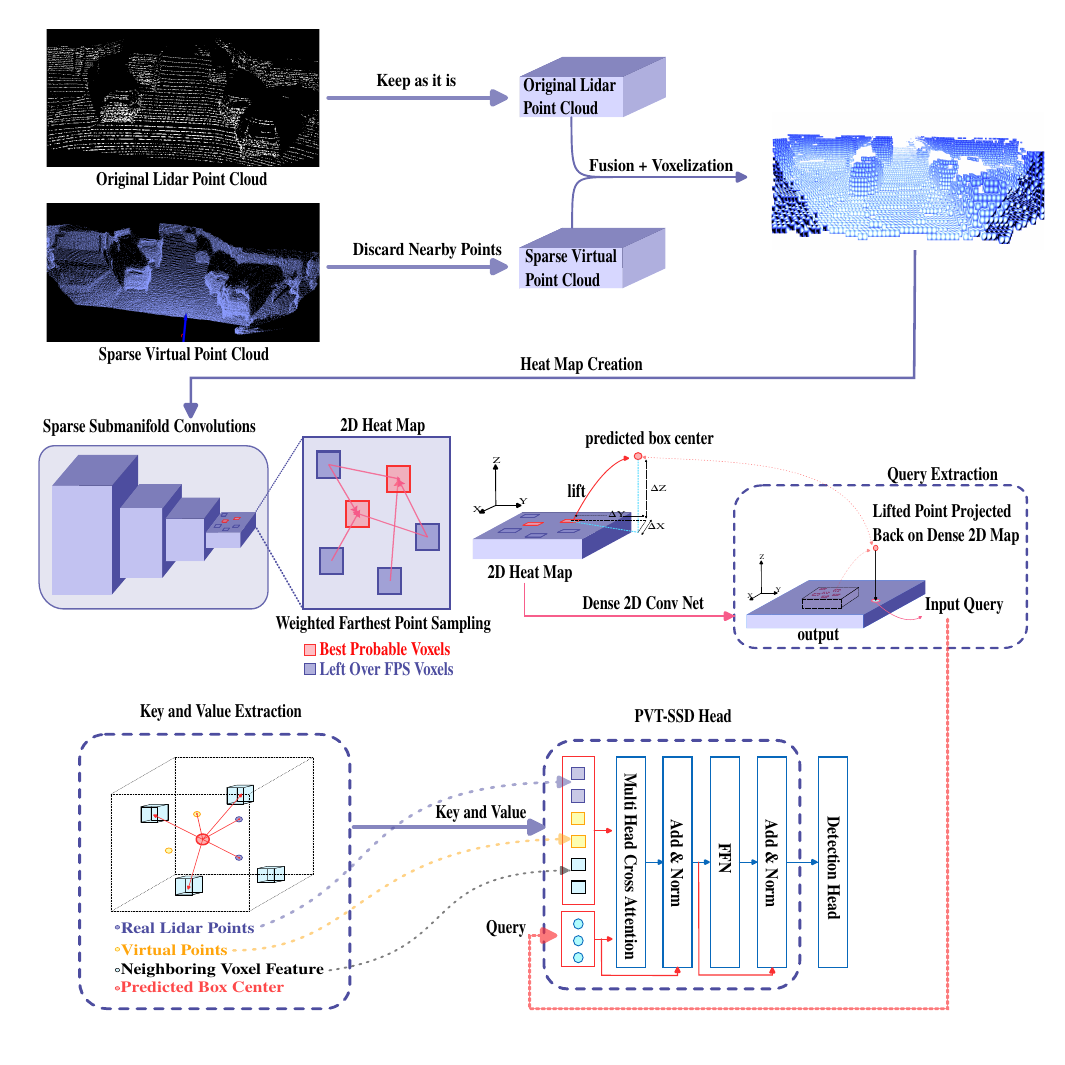}
  \caption{Pipeline: Real and virtual points are fused, voxelized, and encoded by a sparse 3D backbone to produce a Heat Map. Vote-guided query-aware sampling picks up queries from representative points (proto-centers), from the lifted seeds, we extract key and value pairs from nearby points and voxels similar to \cite{b6}. Queries are extracted by picking up the corresponding (x,y) cell features from the densified Heat Map.}
  \label{fig:fig2}
\end{figure*}

Our objective is to develop a transformer model that increases the 3D detection accuracy of distant objects. Here we focus on improving far-field geometry by adding virtual points from RGB images only for \textit{sparse far-field regions}. We also reduce the computation requirements by employing vote-guided sampling and aggregating context over the BEV representation. Figure \ref{fig:fig2} shows our pipeline at a high level.

\subsection{\textbf{Data acquisition and pre-processing}}
We utilize synchronized LiDAR-RGB pairs obtained from a vehicle-mounted multi-sensor platform. The data is partitioned into training-validation and testing sets.
\begin{enumerate}
  \item \textbf{Depth to virtual points:} We first back-project the LiDAR points into the image plane to obtain a depth-map (where each pixel value represents the depth of the projected point). This, along with image samples (and associated LiDAR samples) are center-cropped/padded and fed to image guided depth completion models like \cite{b47,b46} for obtaining completed depth-maps. These are then back-projected into the LiDAR frame giving virtual points. Concretely, given a LiDAR point $\tilde{\mathbf{p}}_L = [x_L, y_L, z_L, 1]^\top$ (in homogeneous coordinate system), its projection in the image plane is given by:
\begin{equation}
    \mathbf{p}_{Img} =
    \begin{bmatrix}
    u \\ v \\ 1
    \end{bmatrix}
    =
    \mathbf{P}_2 \, \mathbf{R}_0 \, \mathbf{T}_{L\to C} \, \tilde{\mathbf{p}}_L,    
    \label{eq:proj}
\end{equation}
where $\mathbf{T}_{L\to C} =
\begin{bmatrix}
\mathbf{R}_{cl} & \mathbf{t}_{cl}\\
\mathbf{0}^\top & 1
\end{bmatrix}$ 
is the rigid LiDAR camera transformation, 
$\mathbf{R}_0$ is the rectification matrix, and 
$\mathbf{P}_2$ is the camera projection matrix with intrinsics $(f_u,f_v,c_u,c_v,t_x,t_y,t_z)$. The depth of each projected point (in camera frame) is obtained as the third coordinate of the transformed LiDAR point, which is given as:
\begin{equation}
    d(u,v) = z_C = [0~0~1] \, (\mathbf{R}_0 \, \mathbf{T}_{L\to C} \, \mathbf{p}_L),    
\end{equation}

This process yields a \emph{sparse} depth map aligned with the corresponding RGB image. The pair (depth map + its RGB Image) is jointly processed by a depth-completion network to estimate a \emph{dense} depth map $\hat{d}(u,v)$. Each pixel $(u,v)$ of this map with its estimated depth $\hat{d}(u,v)$ is then back-projected into 3D using the pinhole camera inverse model:
\begin{equation}
    \mathbf{p}_C =
    \begin{bmatrix}
    x_C \\ y_C \\ z_C
    \end{bmatrix}
    =
    \hat{d}(u,v)
    \begin{bmatrix}
    (u - c_u)/f_u \\ (v - c_v)/f_v \\ 1
    \end{bmatrix},    
\end{equation}
which is transformed into the LiDAR frame, yielding \emph{a virtual point $p_{virt}$}:
\begin{equation}
    \mathbf{p}_{virt} = \mathbf{T}_{C\to L}\,\mathbf{R}_0^{-1}\,
    \begin{bmatrix}
    \mathbf{p}_C \\ 1
    \end{bmatrix}.  
    \label{eq:virt}
\end{equation}

  %yielding virtual points $\{\mathbf{p}_{\text{virt}}\}$ that co-exist with real LiDAR.
  \item \textbf{Range-aware sampling (virtual points only):}
    We partition virtual points into $n_{bins}$ radial bins. We use $n_{bins}{=}2$ during training and $n_{bins}{=}10$ during inference. We retain a fraction $r$ (default $r{=}0.2$) of points from nearby bins (whose distance is $< 60m$ (tunable parameter)) by uniform sampling, and all points from distant bins are retained.
\end{enumerate}

\subsection{\textbf{Point-Cloud Fusion}}
\noindent
As noted in \cite{b2,b3,b4}, voxel features benefit when real and virtual points co-exist in the same spatial bin. Hence, we concatenate real \& virtual points before voxelization (termed as \emph{early fusion} in some texts). Nevertheless, we also provide support for \emph{late fusion}, which involves treating both point clouds separately and fusing them at some later point in the pipeline (other alternative fusion strategies are explored and evaluated in the section \nameref{sec:ablation}).

\subsection{\textbf{Virtual-Point Features}}
We encode each point as an 8-D vector $\mathbf{p}=[x,y,z,I,r,g,b,\tau]$, where $x,y,z$ are Cartesian coordinates, $I$ is LiDAR intensity, $(r,g,b)$ are image colors, and $\tau$ is a modality tag which indicates whether the point is real or virtual. We keep $\tau = 2$ to represent real LiDAR points and $\tau = 1$ for virtual points (projected from the depth map).

\subsection{\textbf{Voxelization and Sparse 3D Backbone}}\label{Heatmap}
% \subsubsection{\textbf{Motivation}}
Transformer\cite{b5} based 3D detectors are highly sensitive to how query tokens are chosen from sparse point clouds.
If we use every non-empty voxel ($n$), then the attention cost and memory will grow quadratically $\mathcal{O}(n^2)$, without much benefit, since most tokens describe background, diluting important object signals.
We therefore begin by \emph{localizing likely object interiors} and then form queries only around those locations.

%\subsubsection{\textbf{Limitations of naïve strategies}}
Some relevant observations suggest that:
\begin{itemize}
    \item Uniform farthest-point sampling (FPS) will strongly bias toward near-field structures (due to high point cloud density) and allocate most queries to uninformative regions (since the percentage of points corresponding to relevant objects is less).
    \item Selecting the top-$K$ cells by foreground logits collapses to a few strong responses and routinely misses small or far-field objects. 
    \item Dense anchor grids increase latency and memory without commensurate recall gains under LiDAR sparsity.

\end{itemize}
All these observations motivate us to choose queries that are \emph{object-centric} (seeded near interiors) as well as \emph{coverage-aware} (spread across the scene and not biased towards near-field regions).

\subsubsection{\textbf{Voxelization}}
We discretize the point cloud into $5$\,cm voxels and,  \emph{for each non-empty voxel} we select one ~\emph{representative point} uniformly at random. This controls memory and mitigates density bias from highly sampled surfaces while preserving the geometry needed for downstream reasoning.

\subsubsection{\textbf{Sparse 3D backbone}}
Following prior work~\cite{b6,b7}, we employ \textbf{sparse submanifold convolution} based backbone, which interleaves submanifold convolutions (propagating features only in occupied sites without dilating sparsity) with occasional strided convolutions (to expand receptive field).
The network produces multi-scale sparse features $\{x_{\mathrm{conv}1},\dots,x_{\mathrm{conv}4}\}$ and a heat map $x_{\mathrm{heat}}$ at 8th stride, which then serves as a compact, geometrically rich canvas (since series of convolutions capture geometry around representative points (peaks) which is impinged on canvas) for subsequent query formation.

\subsubsection{\textbf{Center heatmap}}
Local maxima in this map represent likely object interiors (as shown in \ref{fig:fig1}), these peaks are later lifted to 3D (as explained in the next section) and used as candidate centers (Fig.~\ref{fig:fig2} center), providing a small set of semantically meaningful tokens to the transformer, while avoiding the challenges mentioned in the previous sections.

\subsection{\textbf{Vote-guided, query-aware sampling}}
\subsubsection{\textbf{Heatmap to candidates}}
Given our heatmap $x_{\mathrm{heat}}$ (Sec.~\ref{Heatmap}), we first extract a candidate set $\mathcal{C}$ by non-maximum suppression (NMS) on $x_{\mathrm{heat}}$ with a minimum inter-peak distance. This operation isolates, with spatial uniformity, distinct heat peaks and forms a set of \emph{candidate peaks} ($\mathcal{C}$).
While one may simply select the top-$K$ responses, this will collapse the queries around a few strong modes and will under-represent far-field objects.

\subsubsection{\textbf{Score-modulated coverage}}
To balance confidence and spatial coverage, we sample $K$ seeds from $\mathcal{C}$ using \emph{score-modulated farthest-point sampling} (FPS). 
Concretely, at iteration $t$, let $\mathcal{S}_t$ be the seeds already chosen. 
We select the next seed $c^\star\!\in\!\mathcal{C}\!\setminus\!\mathcal{S}_t$ that maximizes the reweighted distance to the current set:

\begin{equation}
c^\star \;=\; \arg\max_{c\in \mathcal{C}\setminus \mathcal{S}_t}
\ \min_{c'\in \mathcal{S}_t},  \tilde d(c,c'), 
\end{equation}

where 
\begin{equation}
\qquad 
\tilde d(c,c') \,=\, \frac{\|c{-}c'\|_2}{\epsilon + s(c)^\gamma}.
\end{equation}

Here $s(c) = x_{heat}(c)$ which represents the score of a candidate seed and $\gamma\!\ge\!1$ controls how strongly high-score locations repel others. The larger $s(c)$ increases the spacing, preventing mode collapse near a single peak. 
To maintain an exploratory nature, a small fraction of seeds is harvested from the low-score tails of $s$ before continuing the FPS.
This ensures that queries are \emph{object-centric} (biased toward high $s$) and 
\emph{coverage-aware} (well spread across the scene).

\subsubsection{\textbf{Lifting candidates.}}
Following \cite{b6}, each chosen seed at $(u,v)$ is paired with a fixed anchor height $z_a$, to form a 3D anchor $\mathbf{a}=(x,y,z_a)$. 
A vote head regressor then calculates the offset $\Delta=(\Delta x,\Delta y,\Delta z)$ from the Heat Map features at that cell, for producing lifted proto-center (proposed object centers)
\begin{equation}
\hat{\mathbf{p}}=\mathbf{a}+\Delta=(x{+}\Delta x,\ y{+}\Delta y,\ z_a{+}\Delta z).
\label{vote-regression}
\end{equation}

%\subsection{\textbf{Heatmap-Aligned Query Extraction and Context Assembly}}

%\subsubsection{\textbf{Motivation}}
%The process is described in detail below.

\subsubsection{\textbf{Heatmap content alignment}}
After lifting, proto-centers no longer lie on the original voxel lattice. Therefore, simply indexing the \emph{sparse} heatmap tensor at $(x{+}\Delta x,\;y{+}\Delta y)$ often returns empty or aliased features. We therefore propagate (align) the \emph{content} to the projected positions $(x{+}\Delta x,\;y{+}\Delta y)$ on the heatmap grid (via dense 2D convolution layers), so each candidate has a valid, well-conditioned descriptor without paying the cost of 3D convolutions.
We convert the sparse heatmap tensor to a \emph{dense} map by zero-filling empty cells, and refining it with a light residual stack of four $3{\times}3$ convolutions with BN–ReLU
($128{\rightarrow}64{\rightarrow}64{\rightarrow}64{\rightarrow}128$), followed by a skip add to the input.
Each lifted proto-center $\hat{\mathbf{p}}=(x{+}\Delta x,\;y{+}\Delta y,\;z_a{+}\Delta z)$ is projected to the heatmap plane (discarding $z$), and we bilinear-sample its feature from this refined dense map (see fig \ref{fig:fig2} middle).
The sampled vector is concatenated with the seed’s original heatmap features and reduced by a $1{\times}1$ convolution to form the \emph{query token}.
Note that the proto-center \emph{locations} are still determined by \textbf{score-modulated FPS} on the heatmap and remain fixed. The alignment module only provides a well-conditioned query vector for each proto-center. Working on the heatmap grid preserves the translation-equivariant structure used for center prediction while avoiding the memory/latency of 3D kernels.

\subsection{\textbf{Context Assembly and Transformer Head}}
\noindent
\subsubsection{\textbf{Context assembly}}
Answer to the queries generated for each proto-center should naturally lie with the object geometries around it (like voxel and point features), hence, similar to~\cite{b6}, around each lifted proto-center, we gather two complementary key/value tokens:
\begin{itemize}
    \item \emph{Voxel Tokens} from nearby sparse voxels via local KNN (K-Nearest Neighbours), and
    \item \emph{Point Tokens} from fused point cloud (real+virtual points, with their modality tags) via fast range-view indexing (described in \cite{b6}) for fine-grained detail.
\end{itemize}
We then concatenate these two to form a single fused bank, which will serve as keys and values for our transformer head.

\subsubsection{\textbf{Transformer head}}
Let $Q$ be the query tokens of proto-centers and let $(K,V)$ be the concatenated voxel+point banks as described above.
We apply multi-head cross-attention with relative positional bias:
\begin{equation}
\mathrm{Attn}(Q,K,V)=\mathrm{softmax}\!\left(\frac{QK^\top}{\sqrt{d}} + B_{\Delta}\right)V,
\end{equation}

Here $B_{\Delta}$ is the relative positional embedding derived from \cite{b8,b9}.
Each block follows the Pre-LayerNorm (Pre-LN) transformer design in which:
\begin{itemize}
    \item Queries, keys, and values are first normalized before the attention computation
    \item The attention output is added back to the input via a residual connection
    \item A second residual branch applies a position-wise feed-forward network (FFN) consisting of two linear layers with a GELU activation in between.
\end{itemize} 
This configuration improves gradient flow and stabilizes training for deep transformer stacks.
We use $L$ layers and $H$ heads (default $H{=}4$), providing a good compute/quality trade-off in our ablations.
This design keeps queries compact while injecting far-field support via virtual points, and avoids the mode collapse and background overload of naïve top-$K$ or all-voxel querying strategies.

\subsection{\textbf{Detection Head \& Loss}}

Following ~\cite{b6}, the total loss $\mathcal{L}$ is the sum of the following:
\begin{equation}
\mathcal{L}=\lambda_{\text{vote}}\mathcal{L}_{\text{vote}}
+ \lambda_{\text{obj}}\mathcal{L}_{\text{obj}}
+ \lambda_{\text{cls}}\mathcal{L}_{\text{cls}}
+ \lambda_{\text{reg}}\mathcal{L}_{\text{reg}}
+ \lambda_{\text{corner}}\mathcal{L}_{\text{corner}}
\end{equation}
where each component is defined as:
\begin{itemize}
    \item $\mathcal{L}_{vote}$ regresses offsets (as defined in Eqn. (\ref{vote-regression})), to ensure lifted proto-centers correctly align with actual object centers.
    \item $\mathcal{L}_{\text{obj}}$ denotes objectness loss (objectness is a measure of how likely the model thinks that the proposed proto-center is an object center rather than a background point), which is computed using focal loss, to mitigate class imbalance. 
    \item $\mathcal{L}_{\text{cls}}$ represents the classification loss, computed using cross-entropy loss over object categories. 
    \item $\mathcal{L}_{\text{reg}}$ is regression loss applied on 3D bounding box parameters like $(x, y, z, w, l, h, \theta)$. 
    \item At high IOU overlap, $\mathcal{L}_{\text{reg}}$ saturates and becomes less reliable for regressing bounding boxes for harder samples. Hence, we include a $\mathcal{L}_{\text{corner}}$ loss term to supervise on the eight 3D corner points $(x \pm w/2,\, y \pm l/2,\, z \pm h/2)$ between predicted and ground-truth boxes using an L1 loss, which improves geometric alignment of bounding boxes on targets in harder samples.
\end{itemize}
Vote-based regression happens in meters. However, applying the same approach to 3D bounding box parameters, such as $(w, l, h)$, can cause larger objects to dominate the gradients and prevent smaller ones from fitting nicely. As a mitigation strategy, we apply logarithmic scaling to $(w, l, h)$, so that multiplicative variations in object size are transformed into additive differences, stabilizing gradients across objects of different sizes. Further, wrap-arounds, such as $\pm\pi$, introduce discontinuities even when they represent the same orientation. Therefore, to mitigate this, we regress on $(\sin\theta, \cos\theta)$ instead of $\theta$, which solves this problem and also yields smoother gradients.

\section{Experiments And Results}
\begin{figure}
    \centering
    \includegraphics[width=\linewidth]{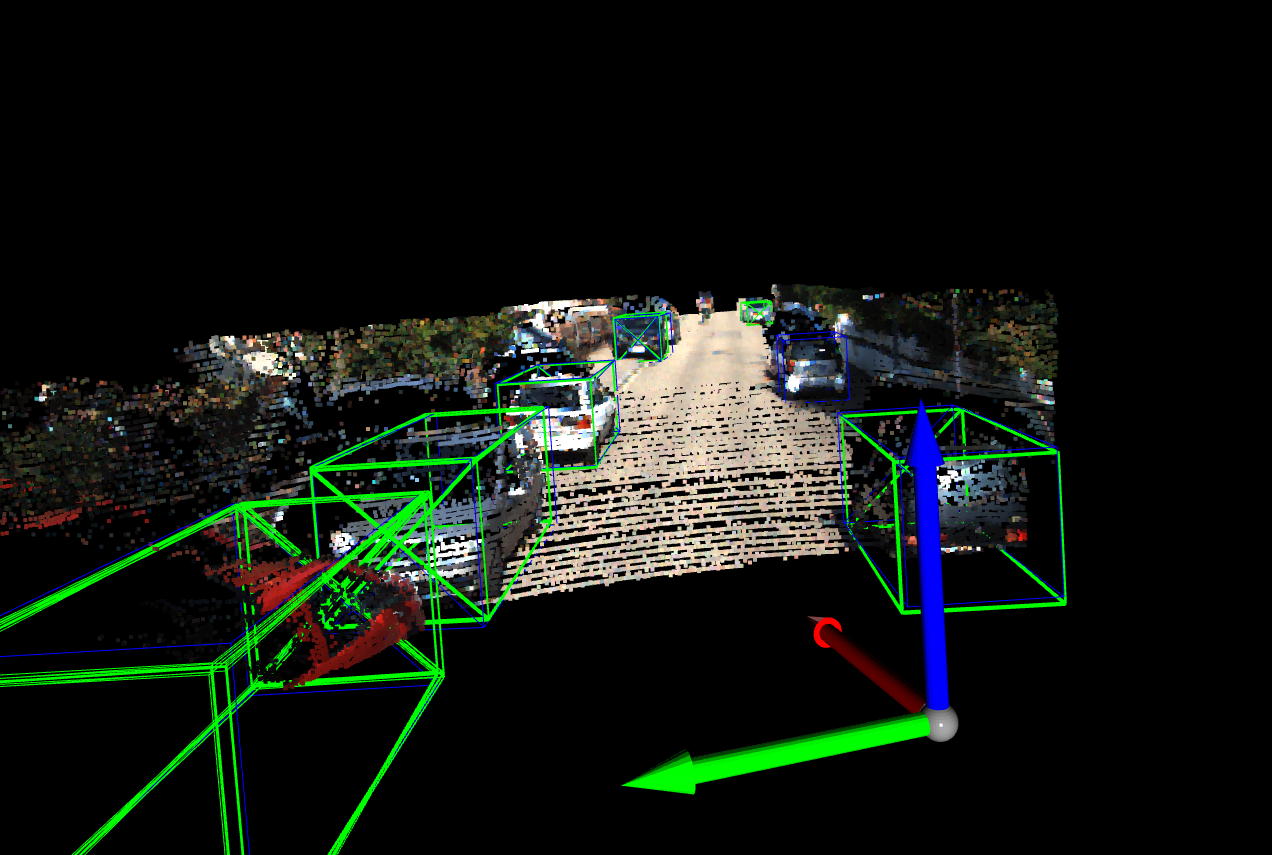}
  \captionof{figure}{Fused real + virtual LiDAR point cloud, where each point is colorized by its corresponding RGB value obtained by back-projection into the image plane using Eqs. (\ref{eq:proj})-(\ref{eq:virt}). The model's 3D detections (cars) are shown as green bounding boxes.}
  \label{fig:fig3}
\end{figure}

\subsection{\textbf{Datasets and Metrics}}
We evaluated our method on the KITTI~\cite{b1} Car split (on the Easy/Moderate/Hard sets) and report the Average Precision (\%AP) for 2D detection, Orientation, 3D detection, and BEV (Bird's-Eye View). We compare our results with those of several other CNN and transformer-based 3D detectors. Results are summarized in Table~\ref{tab:kitti-2d-curated}

\subsection{\textbf{Implementation Details}}
We have used random horizontal flip, global rotations $\pm45^\circ$, global scale $\pm10\%$, and followed ~\cite{b6} for batch sizes. To bias learning toward far-field objects (following \cite{b36}), we drop $80\%$ of voxels within $40$\,m.  

For LiDAR points, we preserve the native $I$, and set $(r,g,b)=(0,0,0)$ (Later on, we set these values to camera projected pixel values in Sec. \ref{sec:paint} but for our final PointViT-V2 model we set this to 0 following~\cite{b36}), and use $\tau{=}2$. 
For virtual points, we set $I{=}0$, and $(r,g,b)$ to the corresponding camera projected pixel values, and use $\tau{=}1$. 
In our PointViT-V2 model, we set $(r,g,b)$ as 0 (for raw LiDAR points) to preserve \emph{geometry only} modalities and prevent over-reliance on image cues.
Restricting RGB to virtual points keeps appearance and geometry disentangled, while adding virtual points increases spatial coverage. Both sets are concatenated before voxelization.

\begin{table}[h]
\caption{KITTI Object Detection and Orientation Estimation Evaluation, run-times are from the leaderboard.}
\label{tab:kitti-2d-curated}
\centering
\begin{threeparttable}
\scriptsize
\setlength{\tabcolsep}{3pt}
\renewcommand{\arraystretch}{1.1}
\begin{tabular*}{\linewidth}{@{\extracolsep{\fill}}lcccc@{}}
\toprule
Method & Mod. & Easy & Hard & Runtime (s) \\
\midrule
    ViKIENet        \cite{b32} & \textbf{97.90} & 98.59 & 92.98 & 0.04 \\
    LVP             \cite{b33} & 97.66 & 98.68 & 92.81 & 0.04 \\
    UDeerPEP        \cite{b34} & 97.39 & 98.40 & \textbf{94.80} & 0.10 \\
    ViKIENet-R      \cite{b32} & 97.08 & 95.78 & 92.11 & 0.06 \\
    VirConv-S       \cite{b36} & 96.46 & 96.99 & 93.74 & 0.09 \\
    GraR-VoI        \cite{b37} & 96.29 & 96.81 & 91.06 & 0.07 \\
    MLF-DET         \cite{b38} & 96.09 & 96.87 & 88.78 & 0.09 \\
    GraR-Po         \cite{b37} & 96.09 & 96.83 & 90.99 & 0.06 \\
    SFD             \cite{b39} & 96.05 & 98.95 & 90.96 & 0.10 \\
    VPFNet          \cite{b40} & 96.04 & 96.63 & 90.99 & 0.06 \\
    VirConv-T       \cite{b36} & 96.01 & 98.64 & 93.12 & 0.09 \\
    TED             \cite{b41} & 95.96 & 96.63 & 93.24 & 0.10 \\
    RDIoU           \cite{b42} & 95.95 & 98.77 & 90.90 & \textbf{0.03} \\
    ACFNet          \cite{b43} & 95.95 & 96.64 & 93.17 & 0.11 \\
    CLOCs           \cite{b44} & 95.93 & 96.77 & 90.93 & 0.10 \\
    GraR-Vo         \cite{b37} & 95.92 & 96.66 & 92.78 & 0.04 \\
    UPIDet          \cite{b45} & 95.85 & 96.25 & 93.17 & 0.11 \\
    PVT-SSD\cite{b6}        & 95.90 & 96.74 & 90.69 & 0.032 \\
\bottomrule
    \textbf{PointViT V1 (Ours)} & 94.04 & \textbf{99.36} & 86.46 & 0.04 \\
    \textbf{PointViT V2 (Ours)} & \textbf{96.56} & 97.04 & 88.97 & 0.04 \\
    \textbf{PointViT P1 (Ours)} & 96.53 & 97.05 & 88.96 & 0.05 \\
\bottomrule
\end{tabular*}
\begin{tablenotes}[flushleft]
\footnotesize
\item Values are AP (\%) for the \emph{2D detection} benchmark (Car class) on the KITTI test server. 
\end{tablenotes}
\end{threeparttable}
\end{table}
\subsection{\textbf{Model Variants}}
\begin{itemize}
    \item \textbf{PVT--SSD (baseline):} The raw PVT--SSD detector without virtual-point fusion.
    \item \textbf{PointViT--V1:} In this variant, we perform early fusion of PE--Net~\cite{b46} virtual points with raw LiDAR before voxelization.
    In this model, we aimed to target moderate and hard samples so the supervision was restricted to Moderate/Hard difficulty (\emph{filter\_by\_difficulty=\{1,2\}}), with a minimum of five LiDAR points required per instance (\emph{filter\_by\_min\_points=\{Car:5, Ped:5, Cyc:5\}}). 
    This design was aimed at preventing trivially easy cases from dominating the losses and encouraging the network to exploit virtual depth cues in sparse regimes.
    \item \textbf{PointViT--V2:} In this variant, we perform early fusion with virtual points extracted from BP-Net\cite {b47}, and prioritize sampling harder examples.
    Training is biased around rarer and more occluded cases, which is achieved via stratified sampling (\emph{SAMPLE\_GROUPS=\{Car:5:0, Car:5:1, Car:5:2, Ped:15, Cyc:15\}}) and enforcing stronger point-count thresholds like (\emph{filter\_by\_min\_points=\{Car:8, Ped:15, Cyc:15\}}), here again we restrict training to Moderate/Hard difficulty (\emph{filter\_by\_difficulty=\{1,2\}}). Dropout is same as in V1 model. This configuration directs capacity toward underrepresented hard cases where virtual points are most beneficial.
    \item \textbf{PointViT-P1:}\label{sec:paint} This one extends our V2 model by introducing RGB augmentation to raw LiDAR points, by back-projecting them onto the image plane, and assigning the corresponding pixel values to those points. Our intention was to provide better visual cues, thereby helping it detect object surface boundaries with better precision. However, accuracy losses suggest that this is of little benefit and often confuses the model further.

\end{itemize}

%%%%%%%%%%%%%%%%%%%%% 100% original checked by originality %%%%%%%%%%%%%%%%%%%%%%%
\begin{figure}[h]
    \centering
    \includegraphics[width=1\linewidth]{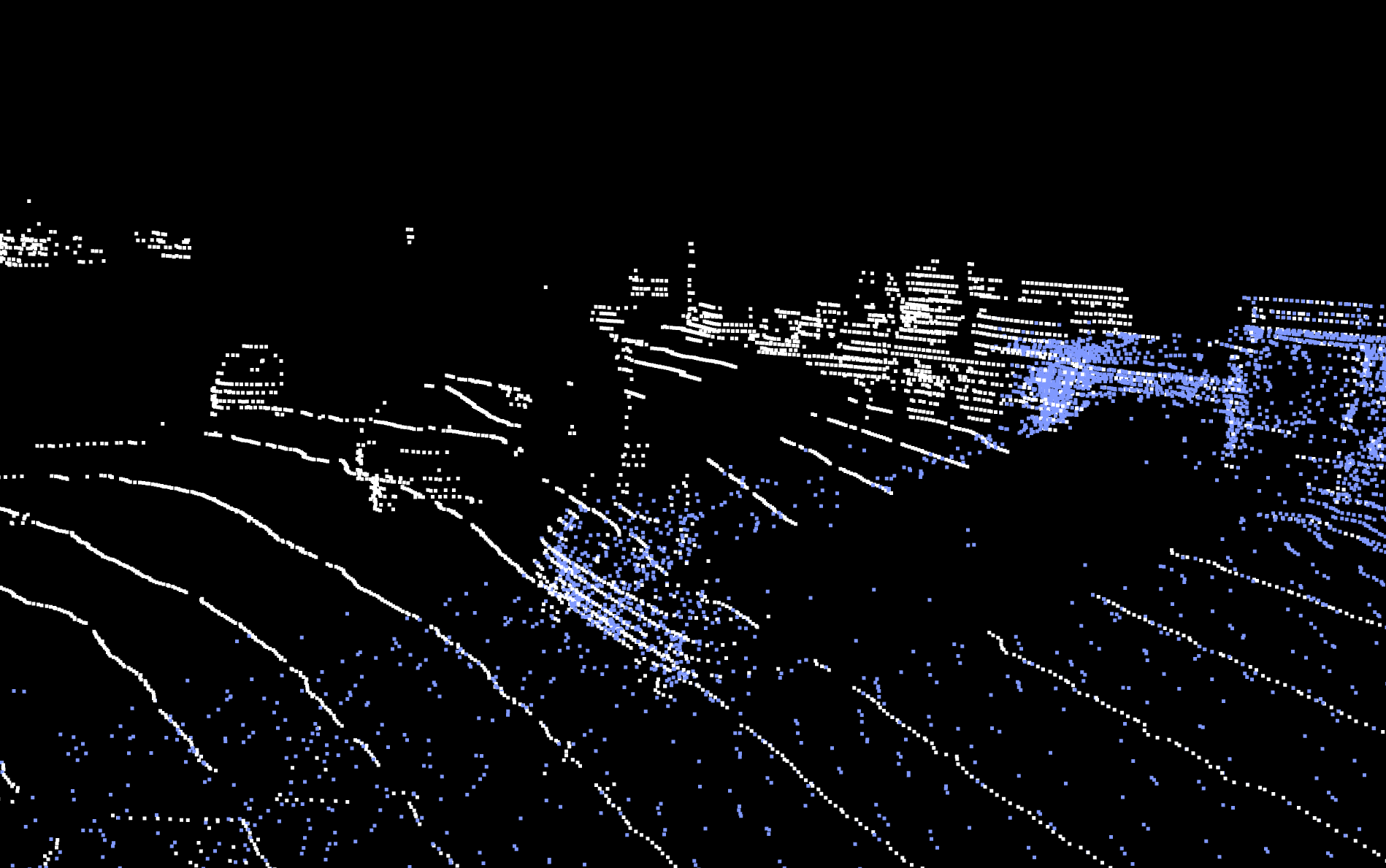}
    \caption{Visualization of fused LiDAR (white) and virtual points (blue). At longer ranges, virtual depth becomes irregular and misaligned with real surfaces, illustrating the non-uniform completion that leads to minor 3D localization drift.}
    \label{fig:depthComplDrift}
\end{figure}

\subsection{\textbf{Discussion}}

    \subsubsection{\textbf{Effect on easy cases}}
    Early fusion results in higher easy set performance (compared with baseline) across all benchmarks, including 3D AP, BEV AP, and 2D/Orientation (Table~\ref{tab:kitti-3d-bev} (3D \& BEV) and Table~\ref{tab:kitti-2d-curated} (2D \& orientation)). This confirms that virtual points provide complementary geometric cues, which enhance \emph{detection} when objects are less occluded.

    \subsubsection{\textbf{Effect on Hard \& Moderate cases}}
    Our detector successfully identifies objects across all difficulty levels, as reflected by strong 2D image and orientation scores (Table~\ref{tab:kitti-2d-curated}), while lagging slightly on 3D and BEV metrics. We attribute this to imperfect box localization at long range: virtual-point generators extrapolate unevenly, producing depth biases that distort geometric consistency, an effect also noted in concurrent studies~\cite{b32}. This phenomenon appears in Fig.~\ref{fig:depthComplDrift}, where far-field virtual points (blue) form fuzzy, uneven bands and occasionally bleed across object boundaries. The resulting non-uniform density of the fused cloud, with over- and under-filled regions, leads to minor centroid drift and footprint mismatch, which explains the drop in 3D and BEV AP for the Moderate and Hard cases.

% Preamble (once): \usepackage{booktabs}

% ============================
% TABLE I — KITTI 3D / BEV AP
% ============================
\begin{table}[t]
\caption{KITTI Car: 3D and BEV AP (\%). Upper rows: test-server results from PVT-SSD\cite{b6}}
\label{tab:kitti-3d-bev}
\centering
\begin{threeparttable}
\scriptsize
\setlength{\tabcolsep}{3pt}
\renewcommand{\arraystretch}{1.1}
\begin{tabular*}{\linewidth}{@{\extracolsep{\fill}}lccccccr@{}}
\toprule
& \multicolumn{3}{c}{3D AP} & \multicolumn{3}{c}{BEV AP} & \multirow{2}{*}{\scriptsize Lat (ms)} \\
\cmidrule(lr){2-4}\cmidrule(lr){5-7}
Method & Easy & Mod. & Hard & Easy & Mod. & Hard & \\
\midrule
\multicolumn{8}{l}{\textit{Two-stage:}}\\
PointRCNN \cite{b11}   & 86.96 & 75.64 & 70.70 & 92.13 & 87.39 & 82.72 & 100 \\
STD \cite{b12}         & 87.95 & 79.71 & 75.09 & 94.74 & 89.19 & 86.42 & 80 \\
PV-RCNN \cite{b13}     & 90.25 & 81.43 & 76.82 & 94.98 & 90.65 & 86.14 & 98 \\
M3DETR \cite{b14}      & 90.28 & 81.73 & 76.96 & 94.41 & 90.37 & 85.98 & -- \\
CT3D \cite{b15}        & 87.83 & 81.77 & 77.16 & 92.36 & 88.83 & 84.07 & 70 \\
PDV \cite{b16}         & 90.43 & 81.86 & 77.36 & 94.56 & 90.48 & 86.23 & 135 \\
PV-RCNN++ \cite{b17}   & 90.14 & 81.88 & 77.15 & 92.06 & 88.74 & 85.97 & 60 \\
EQ-PVRCNN \cite{b18}   & 90.31 & 82.00 & 77.53 & 94.55 & 90.09 & 86.42 & 200 \\
Pyramid-PV \cite{b19}  & 88.39 & 82.08 & 77.49 & 92.19 & 88.84 & 86.21 & 127 \\
VoTr-TSD \cite{b20}    & 89.90 & 82.09 & 79.14 & 94.03 & 90.34 & 86.14 & 216\\
SPG \cite{b21}         & 90.50 & 82.13 & 78.90 & 94.33 & 88.70 & 85.98 & 156 \\
\midrule
\multicolumn{8}{l}{\textit{One-stage:}}\\
SECOND \cite{b22}      & 83.34 & 72.55 & 65.82 & 89.39 & 83.77 & 78.59 & 50 \\
PointPillars \cite{b23}& 82.58 & 74.31 & 68.99 & 90.07 & 86.56 & 82.81 & 24 \\
TANet \cite{b24}       & 84.39 & 75.94 & 68.82 & 91.58 & 86.54 & 81.19 & 35 \\
Point-GNN \cite{b25}   & 88.33 & 79.47 & 72.29 & 93.11 & 89.17 & 83.19 & 643 \\
3DSSD \cite{b26}       & 88.36 & 79.57 & 74.55 & 92.66 & 89.02 & 85.86 & 38 \\
SA-SSD \cite{b27}      & 88.75 & 79.79 & 74.16 & 95.03 & 91.03 & 85.96 & 40 \\
IA-SSD \cite{b28}      & 88.87 & 80.32 & 75.10 & 93.14 & 89.48 & 84.42 & 44 \\
SVGA-Net \cite{b29}    & 87.33 & 80.47 & 75.91 & 92.07 & 89.08 & 85.59 & 62 \\
PVGNet \cite{b30}      & 89.94 & 81.81 & 77.09 & 94.36 & 91.26 & 86.63 & -- \\
SASA \cite{b31}        & 88.76 & 82.16 & 77.16 & 92.87 & 89.51 & 86.35 & 63 \\
PVT-SSD\cite{b6} & 90.65 & \textbf{82.29} & \textbf{76.85} & 95.23 & 91.63 & 86.43 & 32$^\dagger$ \\
\addlinespace[1pt]
\bottomrule
\textbf{PointViT V1(Ours)$^*$} & \textbf{91.16} & 79.93 & 72.51 & \textbf{95.94} & 89.95 & 82.40 & 36$^\dagger$ \\
\textbf{PointViT V2(Ours)$^*$$^*$} & 89.81 & 80.54 & 74.96 & 93.59 & 89.67 & 82.12 & 36$^\dagger$ \\
\textbf{PointViT P1(Ours)$^*$$^*$$^*$} & 89.22 & 79.97 & 72.64 & 93.44 & 89.57 & 82.08 & 50$^\dagger$ \\
\bottomrule
\end{tabular*}
\begin{tablenotes}[flushleft]
\footnotesize
\item $^\dagger$ Latency measured on \textbf{1$\times$ RTX 4090 GPU}, batch size 1.
\item $^*$ Early Fusion + PE-Net used for depth completion.
\item $^*$$^*$ Early Fusion + BP-Net used for depth completion.
\item $^*$$^*$$^*$ Same as V2 but with colored raw LiDAR points.
\end{tablenotes}
\end{threeparttable}
\end{table}

\subsection{\textbf{Ablation Studies}}
\label{sec:ablation}

%\subsubsection{\textbf{\textbf{Protocol}}}
All ablations use the KITTI \textit{Car} split with the baseline parameters and standard augmentations, including near-field dropout set to 80\% within 40m and 0\% beyond that (essentially preserving all virtual points beyond 40m) as in \cite{b36}, and are kept the same unless stated otherwise.
We report Moderate AP (\%) for 2D detection, orientation (AOS), 3D detection, and BEV.
Quantitative results are summarized in Table~\ref{tab:ablation-summary}

\subsubsection{\textbf{\textbf{Experiment 1}: Late BEV Fusion}}
Virtual \& raw LiDAR's point cloud features are encoded separately and fusion is deferred until the BEV stage (hence the name later fusion), where they are fused by simple 1*1 convolution.

\subsubsection{\textbf{\textbf{Experiment 2}: Gated BEV Fusion}}
This one is an extension of Experiment 1, which involves changing the fusion strategy as follows: A mid-network learnable gate mixes virtual point and raw LiDAR's BEV maps, which makes the system preferential to stable signals in case one is over noisy.

\subsubsection{\textbf{\textbf{Experiment 3}: Early Fusion}}
PE--Net is used for generating virtual points which are fused with raw LiDAR points \emph{before voxelization}. Early fusion benefits from dense spatial input for generating voxel features (reduces zero feature regions), resulting in smoother gradients, resulting in better accuracy than later fusion in easy splits, but for far-field sparse LiDAR regions, noisy/misaligned virtual points pollute spatial encoding, causing small centroid /yaw shifts, resulting in a drop of accuracy in Medium and Hard splits. Nevertheless, in concurrent studies like ViKiENet \cite{b32}, Virtual-Instance-to-Real Attention (VIRA) (cross attention-based module) is used for later-fusion followed by feature fusion modules (VIFF) at BEV and RoI levels.

% \subsubsection{\textbf{Experiment 4: Larger Capacity}}
% Heads increased to $8$, FFN width doubled, losses adjusted (\emph{vote\_cls} $1.5{\rightarrow}1.0$, \emph{vote\_reg} $1.2{\rightarrow}2.0$); other settings follow Exp.~3.Results signaled signs of overfitting, suggesting the baseline capacity is better matched to KITTI without stronger regularization or more data.

\subsubsection{\textbf{\textbf{Experiment 4}: Early Fusion (with BP--Net)}}
This is a modification to Experiment 3, where PE-Net is replaced with the better-performing BP-Net for generating virtual points. Furthermore, we use stratified sampling techniques to bias learning toward harder cases (i.e., sampling splits with higher difficulty, keeping higher quotas for Ped/Cyc), stricter point thresholds (Car$\geq$8, Ped/Cyc$\geq$15). Results show Hard-split accuracy improvements, along with higher Moderate 2D/AOS, but do not translate to higher Moderate 3D/BEV under current regularization. 

\begin{table}[t]
\caption{Ablation summary on KITTI Car (Moderate AP, \%). (Test-server results are directly comparable, validation-only entries are diagnostic.)}
\label{tab:ablation-summary}
\centering
\scriptsize
\setlength{\tabcolsep}{3.5pt}
\renewcommand{\arraystretch}{1.1}
\begin{threeparttable}
\begin{tabular*}{\linewidth}{@{\extracolsep{\fill}}lcccc@{}}
\toprule
Method (Exp.\#) & 2D & Orient. & 3D & BEV \\
\midrule
PVT--SSD (baseline)$^{\ddagger}$           & 95.90 & 95.83 & \textbf{82.29} & \textbf{91.63} \\
Exp.~1: Simple Fusion$^\dagger$     & 96.36 & 96.28 & 82.19 & 89.83 \\
Exp.~2: Gated Fusion$^\dagger$ & --    & --    & $\sim$82.8\,(+0.5) & -- \\
Exp.~3: Early Fusion(PE--Net)$^{\ddagger}$ & 94.04 & 93.98 & 79.93 & 89.95 \\
Exp.~4: Early Fusion(BP--Net, hard)$^{\ddagger}$  & \textbf{96.56} & \textbf{96.50} & 80.54 & 89.67 \\
\bottomrule
\end{tabular*}
\begin{tablenotes}[flushleft]
\footnotesize
\item[$^\dagger$] Validation split only, KITTI test submission not retained. Included for diagnostic completeness and not for leaderboard comparison.
\item[$^\ddagger$]Early-fusion results shown are test-server numbers; validation scores for early fusion were higher but are not reported here.

\end{tablenotes}
\end{threeparttable}
\end{table}

\subsection{\textbf{Reproducibility}}
We report our training configurations in this section to encourage reproducibility of our results.
\begin{itemize}
    \item \textbf{Augmentations:} Flip, rotation $\pm45^\circ$, scale $\pm10\%$.
    \item \textbf{Voxel dropout:} $80\%$ within 60m (reduces near field bias).
    \item \textbf{Optimizer:} We use the default scheduler and learning rate as mentioned in ~\cite{b6}.
    \item \textbf{Loss weights:} 
   $\lambda_{cls}=1,\;
    \lambda_{reg}=1,\;
    \lambda_{\text{vote}}=1,\;
    \lambda_{\text{corner}}=3.0.$
\end{itemize}

\section{Conclusion}
In this work, we attempted to address the challenges of far-field sparsity in LiDAR point clouds by augmenting them with virtual points derived from multiple depth-completion methods. We systematically analyzed alternative fusion strategies to minimize long-range misalignment between real and virtual points, thereby improving geometric stability and detection consistency at large distances. The study highlights that while virtual-point fusion can alleviate sparsity, its effectiveness strongly depends on the fusion stage and the uniformity of depth completion across distance. We also observed that virtual point fusion greatly enhances the 2D object detection due to better geometric cues and better far-field 3D detection.


\begin{thebibliography}{00}
\bibitem{b1} Andreas Geiger, Philip Lenz, and Raquel Urtasun. Are we
ready for autonomous driving? The KITTI vision benchmark
suite. In \textit{Proceedings of the IEEE Conference on Computer
Vision and Pattern Recognition, 2012}.
\bibitem{b2}V. A. Sindagi, Y. Zhou and O. Tuzel, "MVX-Net: Multimodal VoxelNet for 3D Object Detection," In \textit{2019 International Conference on Robotics and Automation (ICRA), Montreal, QC, Canada, 2019, pp. 7276-7282}.
\bibitem{b3} S. Vora, A. H. Lang, B. Helou and O. Beijbom, "PointPainting: Sequential Fusion for 3D Object Detection," In \textit{2020 IEEE/CVF Conference on Computer Vision and Pattern Recognition (CVPR), Seattle, WA, USA, 2020, pp. 4603-4611.}
\bibitem{b4} C. Wang, C. Ma, M. Zhu and X. Yang, "PointAugmenting: Cross-Modal Augmentation for 3D Object Detection," In \textit{ 2021 IEEE/CVF Conference on Computer Vision and Pattern Recognition (CVPR), Nashville, TN, USA, 2021, pp. 11789-11798.}
\bibitem{b5} Ashish Vaswani, Noam Shazeer, Niki Parmar, Jakob Uszkoreit, Llion Jones, Aidan N. Gomez, Lukasz Kaiser, and Illia
Polosukhin. Attention is all you need. In \textit{Advances in Neural
Information Processing Systems, 2017}.
\bibitem{b6} H. Yang et al., "PVT-SSD: Single-Stage 3D Object Detector with Point-Voxel Transformer," In \textit{2023 IEEE/CVF Conference on Computer Vision and Pattern Recognition (CVPR), Vancouver, BC, Canada, 2023, pp. 13476-13487.}

\bibitem{b7} B. Graham, M. Engelcke and L. v. d. Maaten, "3D Semantic Segmentation with Submanifold Sparse Convolutional Networks," In \textit{2018 IEEE/CVF Conference on Computer Vision and Pattern Recognition, Salt Lake City, UT, USA, 2018, pp. 9224-9232.}

\bibitem{b8} Kan Wu, Houwen Peng, Minghao Chen, Jianlong Fu, and
Hongyang Chao. Rethinking and improving relative position
encoding for vision transformer. In \textit{Proceedings of the IEEE
International Conference on Computer Vision, 2021.}

\bibitem{b9} Zetong Yang, Li Jiang, Yanan Sun, Bernt Schiele, and Jiaya Jia. A unified query-based paradigm for point cloud understanding. In \textit{Proceedings of the IEEE Conference on Computer Vision and Pattern Recognition, 2022.}

\bibitem{b10} C. R. Qi, O. Litany, K. He and L. Guibas, "Deep Hough Voting for 3D Object Detection in Point Clouds," In \textit{2019 IEEE/CVF International Conference on Computer Vision (ICCV), Seoul, Korea (South), 2019, pp. 9276-9285.}

\bibitem{b11} S. Shi, X. Wang, and H. Li,
“PointRCNN: 3D Object Proposal Generation and Detection from Point Cloud,”
in \textit{Proc. IEEE/CVF Conf. Comput. Vis. Pattern Recognit. (CVPR)}, 2019, pp. 770--779.

\bibitem{b12} Z. Yang, Y. Sun, S. Liu, and J. Jia,
“STD: Sparse-to-Dense 3D Object Detector for Point Cloud,”
in \textit{Proc. IEEE/CVF Int. Conf. Comput. Vis. (ICCV)}, 2019, pp. 1951--1960.

\bibitem{b13} S. Shi, C. Guo, L. Jiang, Z. Wang, J. Shi, X. Wang, and H. Li,
“PV-RCNN: Point-Voxel Feature Set Abstraction for 3D Object Detection,”
in \textit{Proc. IEEE/CVF Conf. Comput. Vis. Pattern Recognit. (CVPR)}, 2020, pp. 10529--10538.

\bibitem{b14} T. Guan, J. Wang, S. Lan, R. Chandra, Z. Wu, L. S. Davis, and D. Manocha,
“M3DeTR: Multi-representation, Multi-scale, Mutual-relation 3D Object Detection with Transformers,”
in \textit{Proc. IEEE/CVF Winter Conf. Appl. Comput. Vis. (WACV)}, 2022, pp. 2904--2913.

\bibitem{b15} H. Sheng, S. Cai, Y. Liu, B. Deng, J. Huang, X.-S. Hua, and M.-J. Zhao,
“Improving 3D Object Detection with Channel-wise Transformer,”
in \textit{Proc. IEEE/CVF Int. Conf. Comput. Vis. (ICCV)}, 2021, pp. 2743--2752.

\bibitem{b16} J. S. K. Hu, T. Kuai, and S. L. Waslander,
“Point Density-Aware Voxels for LiDAR 3D Object Detection,”
in \textit{Proc. IEEE/CVF Conf. Comput. Vis. Pattern Recognit. (CVPR)}, 2022, pp. 8469--8478.

\bibitem{b17} S. Shi, L. Jiang, J. Deng, Z. Wang, C. Guo, J. Shi, X. Wang, and H. Li,
“PV-RCNN++: Point-Voxel Feature Set Abstraction With Local Vector Representation for 3D Object Detection,”
\textit{Int. J. Comput. Vis. (IJCV)}, vol. 131, pp. 2510--2533, 2023. (Early version: arXiv:2102.00463.)

\bibitem{b18} Z. Yang, L. Jiang, Y. Sun, B. Schiele, and J. Jia,
“A Unified Query-based Paradigm for Point Cloud Understanding,”
in \textit{Proc. IEEE/CVF Conf. Comput. Vis. Pattern Recognit. (CVPR)}, 2022.

\bibitem{b19} J. Mao, M. Niu, H. Bai, X. Liang, H. Xu, and C. Xu,
“Pyramid R-CNN: Towards Better Performance and Adaptability for 3D Object Detection,”
in \textit{Proc. IEEE/CVF Int. Conf. Comput. Vis. (ICCV)}, 2021, pp. 2723--2732.

\bibitem{b20} J. Mao, Y. Xue, M. Niu, H. Bai, J. Feng, X. Liang, H. Xu, and C. Xu,
“Voxel Transformer for 3D Object Detection,”
in \textit{Proc. IEEE/CVF Int. Conf. Comput. Vis. (ICCV)}, 2021, pp. 3164--3173. (VoTr-TSD variant.)

\bibitem{b21} Q. Xu, Y. Zhou, W. Wang, C. R. Qi, and D. Anguelov,
“SPG: Unsupervised Domain Adaptation for 3D Object Detection via Semantic Point Generation,”
in \textit{Proc. IEEE/CVF Int. Conf. Comput. Vis. (ICCV)}, 2021, pp. 15446--15456.

\bibitem{b22} Y. Yan, Y. Mao, and B. Li,
“SECOND: Sparsely Embedded Convolutional Detection,”
\textit{Sensors}, vol. 18, no. 10, p. 3337, 2018.

\bibitem{b23} A. H. Lang, S. Vora, H. Caesar, L. Zhou, J. Yang, and O. Beijbom,
“PointPillars: Fast Encoders for Object Detection From Point Clouds,”
in \textit{Proc. IEEE/CVF Conf. Comput. Vis. Pattern Recognit. (CVPR)}, 2019, pp. 12697--12705.

\bibitem{b24} Z. Liu, X. Zhao, T. Huang, R. Hu, Y. Zhou, and X. Bai,
“TANet: Robust 3D Object Detection from Point Clouds with Triple Attention,”
in \textit{Proc. AAAI Conf. Artif. Intell. (AAAI)}, vol. 34, no. 7, 2020, pp. 11677--11684.

\bibitem{b25} W. Shi and R. Rajkumar,
“Point-GNN: Graph Neural Network for 3D Object Detection in a Point Cloud,”
in \textit{Proc. IEEE/CVF Conf. Comput. Vis. Pattern Recognit. (CVPR)}, 2020, pp. 1711--1719.

\bibitem{b26} Z. Yang, Y. Sun, S. Liu, and J. Jia,
“3DSSD: Point-Based 3D Single Stage Object Detector,”
in \textit{Proc. IEEE/CVF Conf. Comput. Vis. Pattern Recognit. (CVPR)}, 2020, pp. 11040--11048.

\bibitem{b27} C. He, H. Zeng, J. Huang, X.-S. Hua, and L. Zhang,
“Structure Aware Single-Stage 3D Object Detection From Point Cloud,”
in \textit{Proc. IEEE/CVF Conf. Comput. Vis. Pattern Recognit. (CVPR)}, 2020, pp. 11873--11882.

\bibitem{b28} Y. Zhang, J. Zhang, J. Sun, and B. Schiele,
“IA-SSD: Instance-Aware Single-Stage 3D Object Detection,”
in \textit{Proc. IEEE/CVF Conf. Comput. Vis. Pattern Recognit. (CVPR)}, 2022, pp. 11086--11095.

\bibitem{b29} Q. He, Z. Wang, H. Zeng, Y. Zeng, and Y. Liu,
“SVGA-Net: Sparse Voxel-Graph Attention Network for 3D Object Detection from Point Clouds,”
in \textit{Proc. AAAI Conf. Artif. Intell. (AAAI)}, 2022, pp. 870--878.

\bibitem{b30} Z. Miao, Z. Wei, Y. Liu, S. Chen, and T. Zhang,
“PVGNet: A Bottom-Up One-Stage 3D Object Detector with Integrated Multi-Level Features,”
in \textit{Proc. IEEE/CVF Conf. Comput. Vis. Pattern Recognit. (CVPR) Workshops}, 2021, pp. 3059--3068.

\bibitem{b31} C. Chen, Z. Chen, J. Zhang, and D. Tao,
“SASA: Semantics-Augmented Set Abstraction for Point-based 3D Object Detection,”
in \textit{Proc. AAAI Conf. Artif. Intell. (AAAI)}, vol. 36, 2022, pp. 221--229.

% b32 — ViKIENet (and ViKIENet-R should reuse this key)
\bibitem{b32}
Z. Yu, B. Qiu, and A. Khong, 
“ViKIENet: Towards Efficient 3D Object Detection with Virtual Key Instance Enhanced Network,” 
in \textit{Proc. IEEE/CVF Conf. Comput. Vis. Pattern Recognit. (CVPR)}, 2025.

% b33 — LVP
\bibitem{b33}
Y. Chen, G. Cai, Z. Song, Z. Liu, B. Zeng, J. Li, and Z. Wang,
“LVP: Leverage Virtual Points in Multi-Modal Early Fusion for 3D Object Detection,”
\textit{IEEE Trans. Geosci. Remote Sens.}, 2024.

% b34 — UDeerPEP (PeP)
\bibitem{b34}
Z. Dong, H. Ji, X. Huang, W. Zhang, X. Zhan, and J. Chen,
“PeP: A Point Enhanced Painting Method for Unified Point Cloud Tasks,”
\textit{arXiv preprint}, 2023.

% b35 — PointViT V2 (Ours) — placeholder; update title/authors/arXiv as needed
\bibitem{b35}
[Authors Omitted],
“PointViT V2: Early-Fusion Virtual Points for 3D Object Detection,”
\textit{arXiv preprint arXiv:XXXX.XXXXX}, 2025.

% b36 — VirConv (use for both VirConv-S and VirConv-T)
\bibitem{b36}
H. Wu, C. Wen, S. Shi, and C. Wang,
“Virtual Sparse Convolution for Multimodal 3D Object Detection,”
in \textit{Proc. IEEE/CVF Conf. Comput. Vis. Pattern Recognit. (CVPR)}, 2023.

% b37 — GraR (use for GraR-VoI, GraR-Po, GraR-Vo)
\bibitem{b37}
H. Yang, Z. Liu, X. Wu, W. Wang, W. Qian, X. He, and D. Cai,
“Graph R-CNN: Towards Accurate 3D Object Detection with Semantic-Decorated Local Graph,”
in \textit{Proc. Eur. Conf. Comput. Vis. (ECCV)}, 2022.

% b38 — MLF-DET
\bibitem{b38}
Z. Lin, Y. Shen, S. Zhou, S. Chen, and N. Zheng,
“MLF-DET: Multi-Level Fusion for Cross-Modal 3D Object Detection,”
in \textit{Proc. Int. Conf. Artif. Neural Netw. (ICANN)}, 2023.

% b39 — SFD
\bibitem{b39}
X. Wu, L. Peng, H. Yang, L. Xie, C. Huang, C. Deng, H. Liu, and D. Cai,
“Sparse Fuse Dense: Towards High-Quality 3D Detection with Depth Completion,”
in \textit{Proc. IEEE/CVF Conf. Comput. Vis. Pattern Recognit. (CVPR)}, 2022.

% b40 — VPFNet
\bibitem{b40}
H. Zhu, J. Deng, Y. Zhang, J. Ji, Q. Mao, H. Li, and Y. Zhang,
“VPFNet: Improving 3D Object Detection with Virtual Point-based LiDAR and Stereo Data Fusion,”
\textit{IEEE Trans. Multimedia}, 2022.

% b41 — TED
\bibitem{b41}
H. Wu, C. Wen, W. Li, R. Yang, and C. Wang,
“Transformation-Equivariant 3D Object Detection for Autonomous Driving,”
in \textit{Proc. AAAI Conf. Artif. Intell. (AAAI)}, 2023.

% b42 — RDIoU
\bibitem{b42}
H. Sheng, S. Cai, N. Zhao, B. Deng, J. Huang, X. Hua, M. Zhao, and G. Lee,
“Rethinking IoU-based Optimization for Single-Stage 3D Object Detection,”
in \textit{Proc. Eur. Conf. Comput. Vis. (ECCV)}, 2022.

% b43 — ACF-Net
\bibitem{b43}
Y. Tian, X. Zhang, X. Wang, J. Xu, J. Wang, R. Ai, W. Gu, and W. Ding,
“ACF-Net: Asymmetric Cascade Fusion for 3D Detection with LiDAR Point Clouds and Images,”
\textit{IEEE Trans. Intell. Vehicles}, 2023.

% b44 — CLOCs
\bibitem{b44}
S. Pang, D. Morris, and H. Radha,
“CLOCs: Camera–LiDAR Object Candidates Fusion for 3D Object Detection,”
in \textit{Proc. IEEE/RSJ Int. Conf. Intell. Robots Syst. (IROS)}, 2020.

% b45 — UPIDet
\bibitem{b45}
Y. Zhang, Q. Zhang, J. Hou, Y. Yuan, and G. Xing,
“Unleash the Potential of Image Branch for Cross-modal 3D Object Detection,”
in \textit{Advances in Neural Information Processing Systems (NeurIPS)}, 2023.

\bibitem{b46} Mu Hu, Shuling Wang, Bin Li, Shiyu Ning, Li Fan, and
Xiaojin Gong. Penet: Towards precise and efficient image
guided depth completion. In \textit{International Conference
on Robotics and Automation (ICRA), pages 13656–13662,
2021}

\bibitem{b47} J. Tang, F. -P. Tian, B. An, J. Li and P. Tan, "Bilateral Propagation Network for Depth Completion," 2024 IEEE/CVF Conference on Computer Vision and Pattern Recognition (CVPR), Seattle, WA, USA, 2024, pp. 9763-9772


\end{thebibliography}
\end{document}